\documentclass[letterpaper, 10 pt, conference]{ieeeconf}  

\IEEEoverridecommandlockouts                              

\PassOptionsToPackage{usenames,dvipsnames}{color}
\usepackage{cite}
\usepackage{amsmath}
\usepackage{pifont}
\newcommand{\cmark}{\ding{51}}%
\newcommand{\xmark}{\ding{55}}%
\usepackage{algorithm}
\usepackage{algpseudocode}
\usepackage{amsfonts}
\usepackage{pgfplots}
\pgfplotsset{compat=1.17}
\usepackage{hyperref}
\usepackage{booktabs}
\usepgfplotslibrary{groupplots} 
\usepackage[dvipsnames]{xcolor}

\title{\LARGE \bf
Whole-Body Model-Predictive Control of Legged Robots with MuJoCo
}

\author{
John Z. Zhang$^{1}$, Taylor A. Howell$^{2}$, Zeji Yi$^{3}$, Chaoyi Pan$^{3}$, Guanya Shi$^{4}$, Guannan Qu$^{3}$,\\ Tom Erez$^{2}$, Yuval Tassa$^{2}$, and Zachary Manchester$^{1}$
\thanks{$^{1}$ John Z. Zhang and Zachary Manchester are with the Department Aeronautics and Astronautics, Massachusetts Institute of Technology.}
\thanks{$^{2}$Taylor A. Howell, Tom Erez, and Yuval Tassa are with Google DeepMind.}
\thanks{$^{3}$Zeji Yi, Chaoyi Pan, and Guannan Qu are with the Department of Electrical and Computer Engineering, Carnegie Mellon University.}
\thanks{$^{4}$Guanya Shi is with the Robotics Institute, Carnegie Mellon University.
        }%
\thanks{This work was completed at Carnegie Mellon University.}
\thanks{Corresponding email: {\tt\small jzhang3@mit.edu}}
}

\begin{document}

\maketitle
\thispagestyle{empty}
\pagestyle{empty}

\begin{abstract}
    We demonstrate the surprising real-world effectiveness of a very simple approach to whole-body model-predictive control (MPC) of quadruped and humanoid robots: the iterative linear-quadratic regulator (iLQR) algorithm with MuJoCo dynamics and finite-difference approximated derivatives. Building upon the previous success of model-based behavior synthesis and control of locomotion and manipulation tasks with MuJoCo in simulation, we show that these policies can easily generalize to the real world with few sim-to-real considerations. Our baseline method achieves real-time MPC while leveraging \emph{whole-body} dynamics and collision detection on a variety of hardware experiments, including dynamic quadruped locomotion, a quadruped walking on two legs, and full-sized humanoid bipedal locomotion. Additionally, our GUI system enables users to interactively update robot behavior in real-time on the robot hardware, making task-specific parameter tuning easy and intuitive. 
    Our code and experiment videos are available online at:\href{https://johnzhang3.github.io/mujoco_ilqr/}{https://johnzhang3.github.io/mujoco\_ilqr}.
\end{abstract}
\section{Introduction}

Enabling legged robots to achieve human and animal-level agility has been a decades-long challenge for robotics researchers. In addition to the challenges faced by other non-legged mobile robots (e.g., drones, autonomous vehicles, etc.), legged systems are generally high-dimensional and must effectively reason about making and breaking contact with the world. 
Advancements in model-based control~\cite{Deits2023Robot, grandia2023perceptive} and reinforcement learning (RL)~\cite{Lee2024Learning, Hoeller2024ANYmal} methods have unlocked tremendous in-the-wild legged robot capabilities over the last $10$-$15$ years.  Over the same period, robotics simulation has seen significant growth in terms of physical accuracy~\cite{drake, todorov2012mujoco, howelllecleach2022}, differentiability~\cite{howelllecleach2022, warp2022}, and parallelization performance~\cite{todorov2012mujoco, warp2022, rudin2022learningwalkminutesusing}. Thanks to these advances in simulation technologies combined with incredible tools supported by the broader machine learning community such as PyTorch~\cite{paszke2019pytorch} and JAX~\cite{jax2018github}, sim-to-real RL has enjoyed accelerated progress and become the standard approach for solving challenging problems like humanoid whole-body control. Interestingly, in comparison, model-based control researchers have generally favored novel, custom implementations of robot models~\cite{howelllecleach2022,carpentier2024compliantrigidcontactsimulation} and optimization solvers~\cite{Howell2019ALTRO, cleach2024fast, jallet2023proxddp}, in part due to the online computation requirements of model-predictive control (MPC) paradigm, making these works relatively more difficult to reproduce which, so far, has resulted in slower community adoption.

\begin{figure}
    \centering
    \includegraphics[trim={1390 425pt 260 275pt}, clip, width=0.24\linewidth]{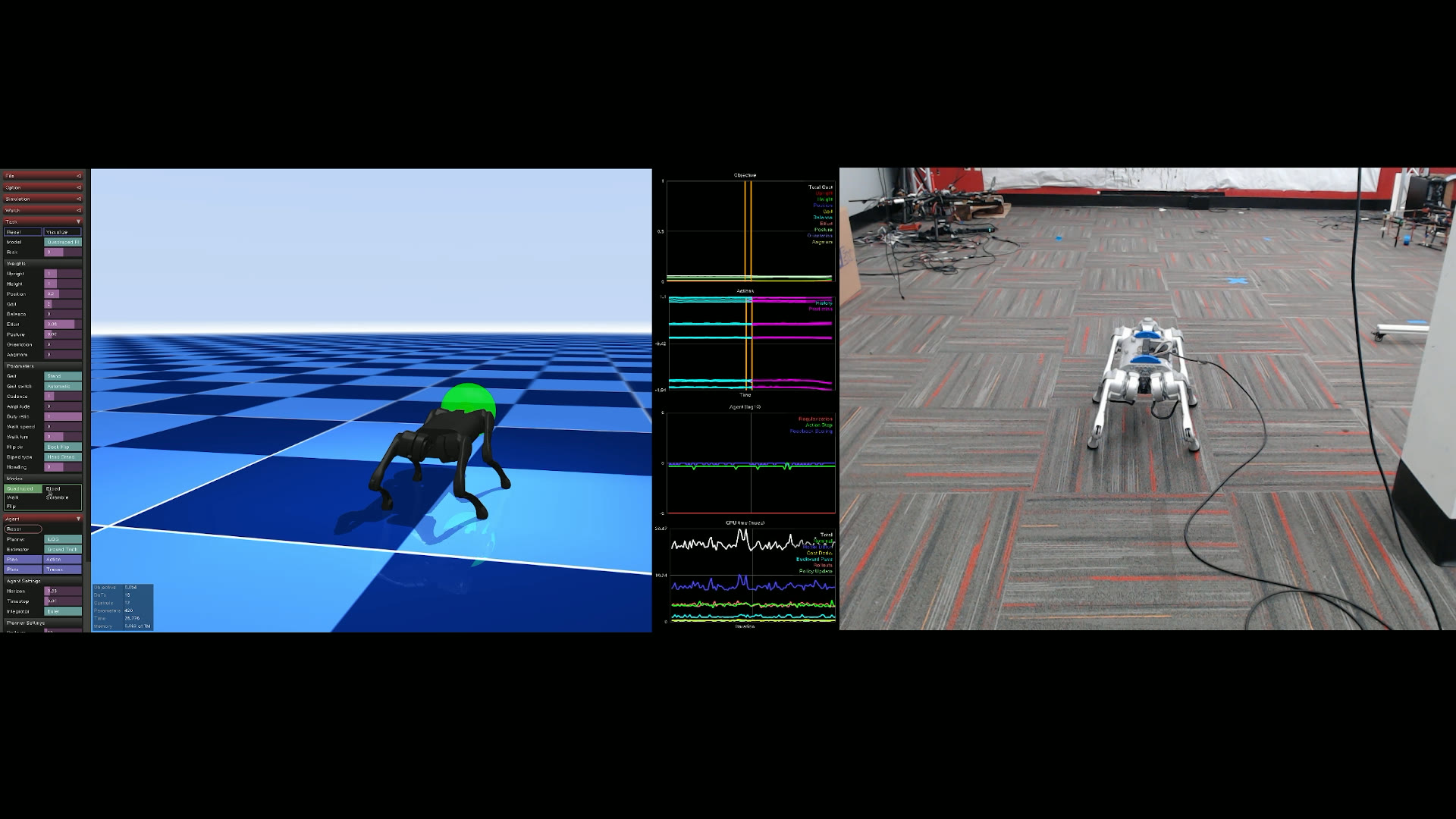}
    \hfill
    \includegraphics[trim={1390 425pt 260 275pt}, clip, width=0.24\linewidth]{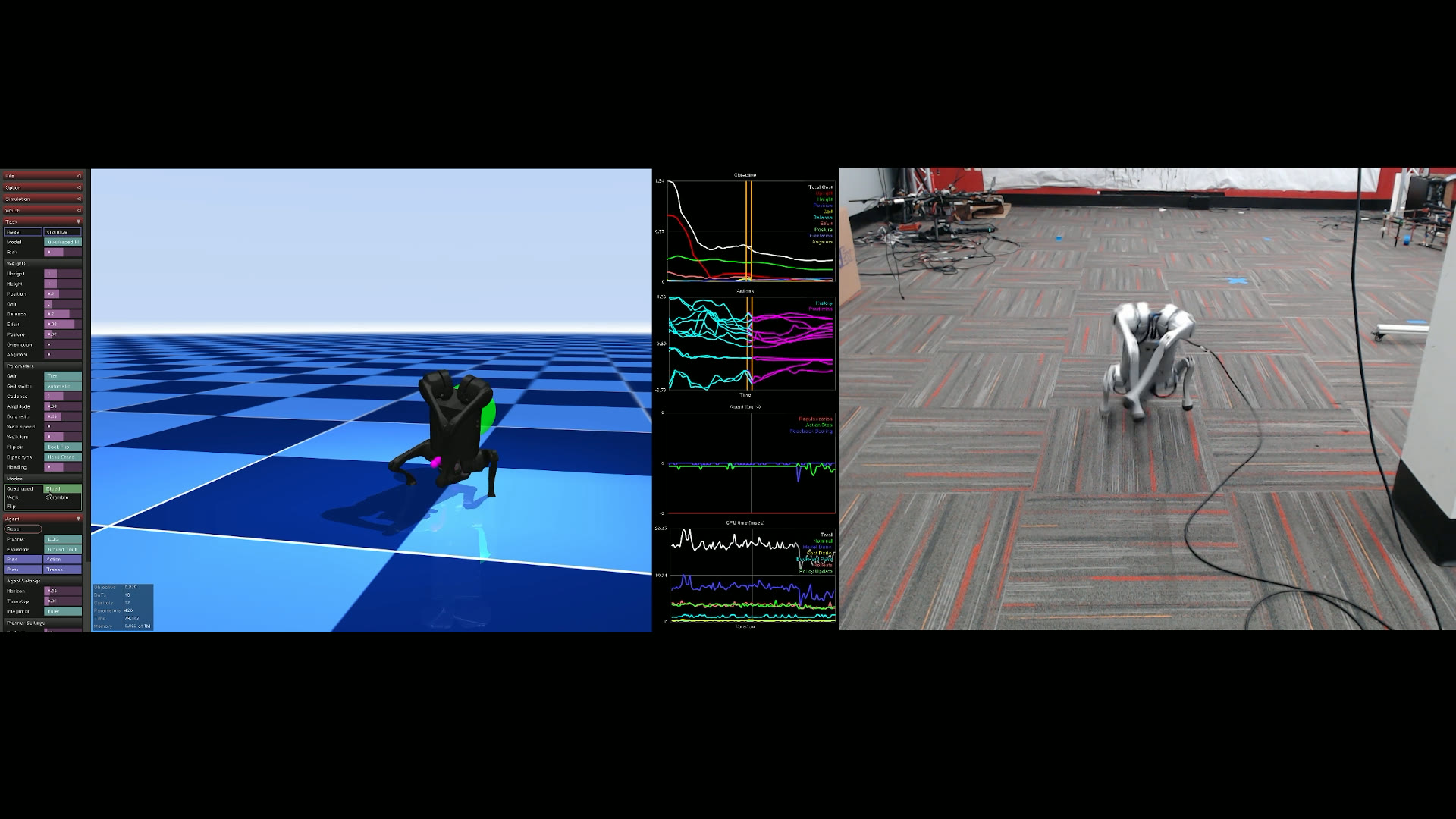}
    \hfill
    \includegraphics[trim={1390 425pt 260 275pt}, clip, width=0.24\linewidth]{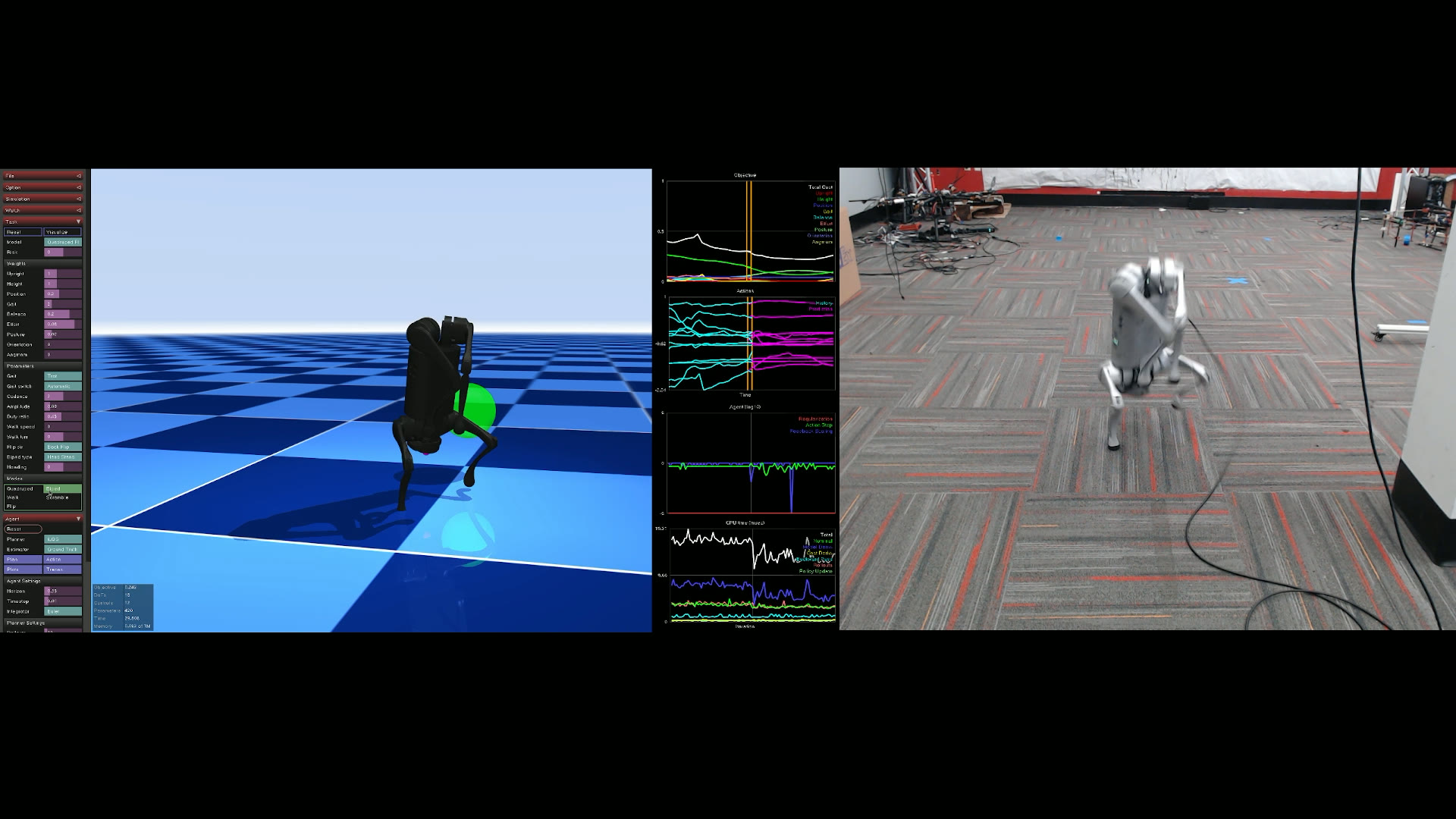}
    \hfill
    \includegraphics[trim={1390 425pt 260 275pt}, clip, width=0.24\linewidth]{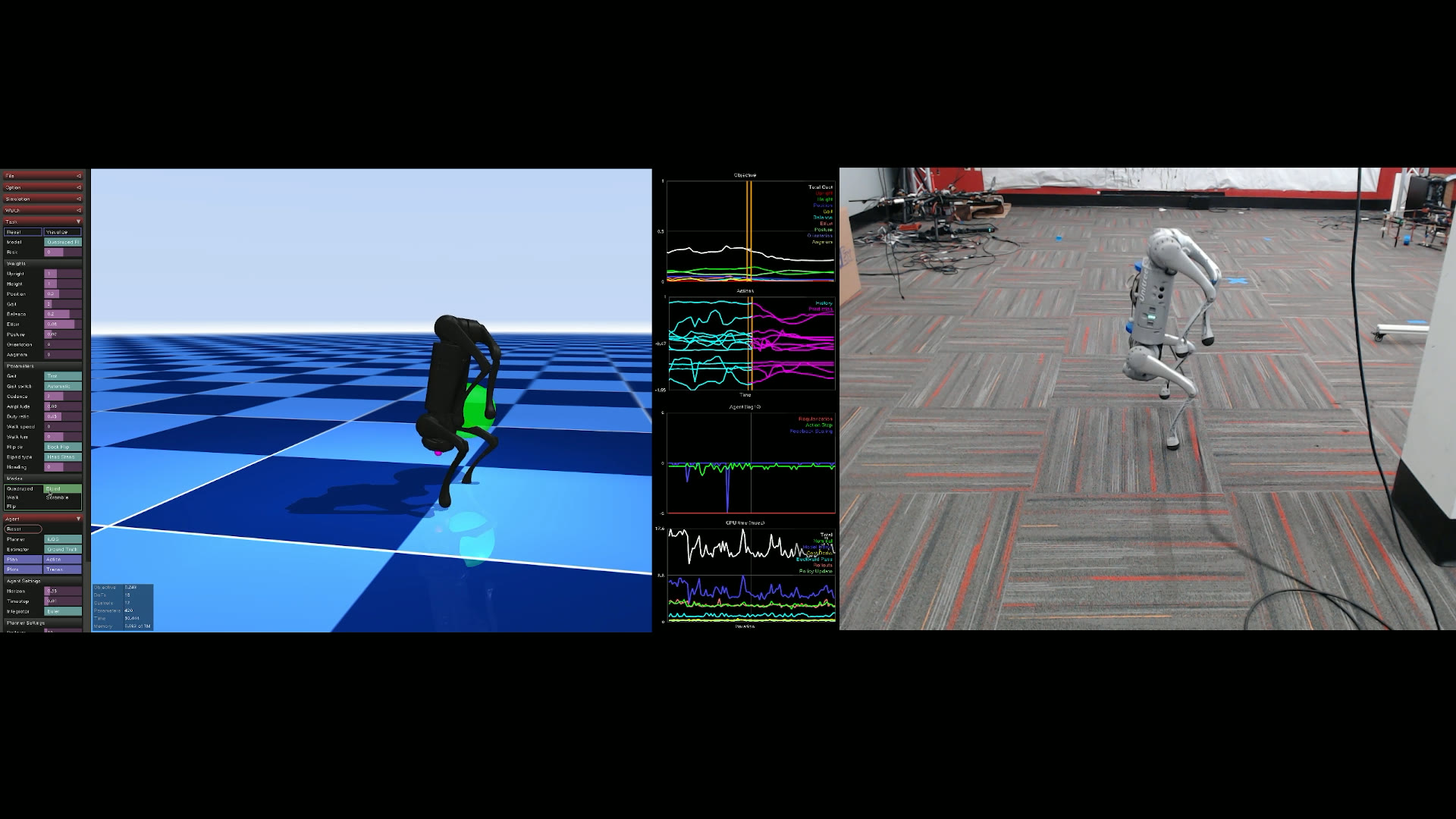}\\
    \vspace{1mm}
    \includegraphics[trim={1350 300pt 300 400pt}, clip, width=0.24\linewidth]{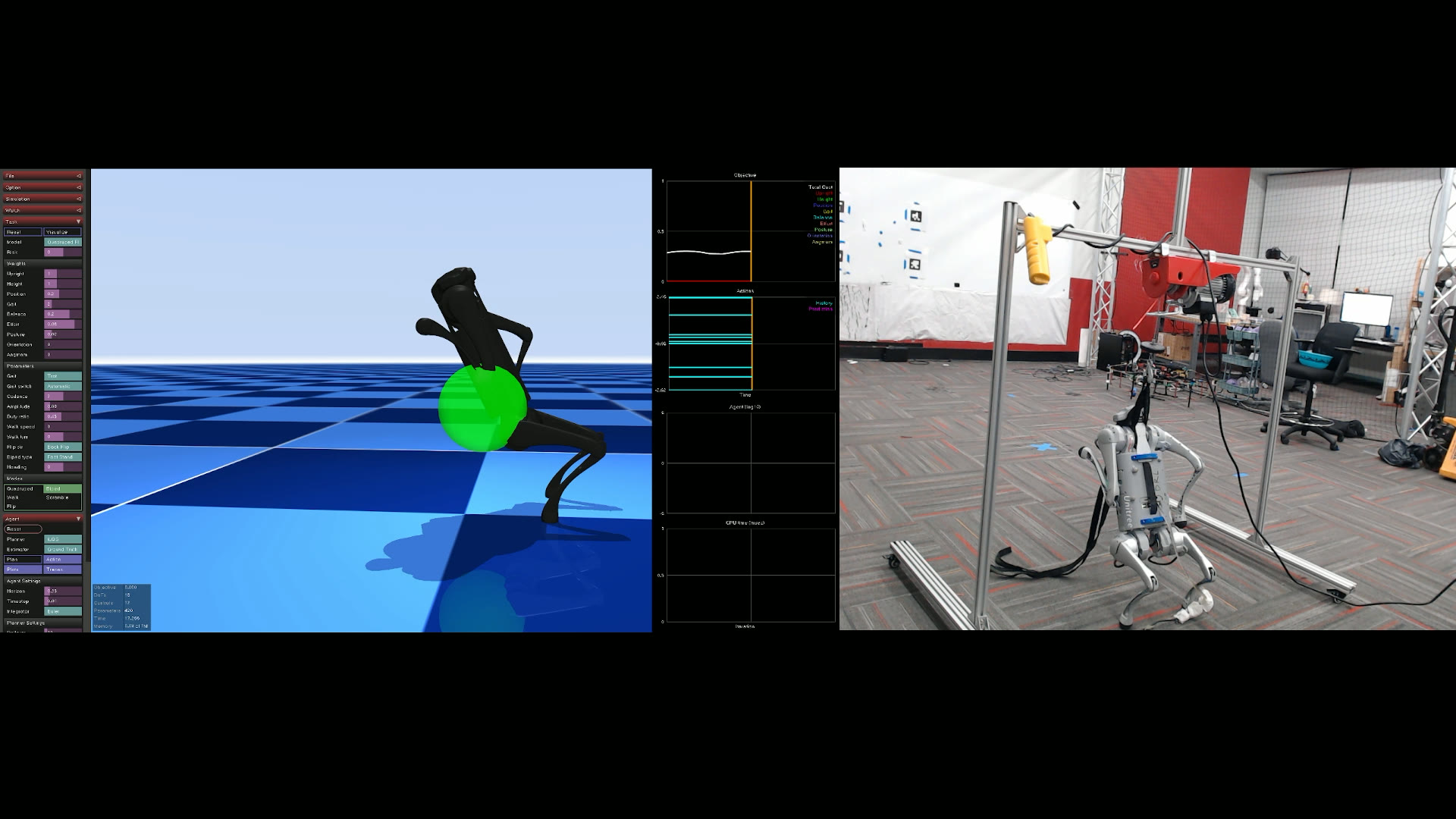}
    \hfill
    \includegraphics[trim={1350 300pt 300 400pt}, clip, width=0.24\linewidth]{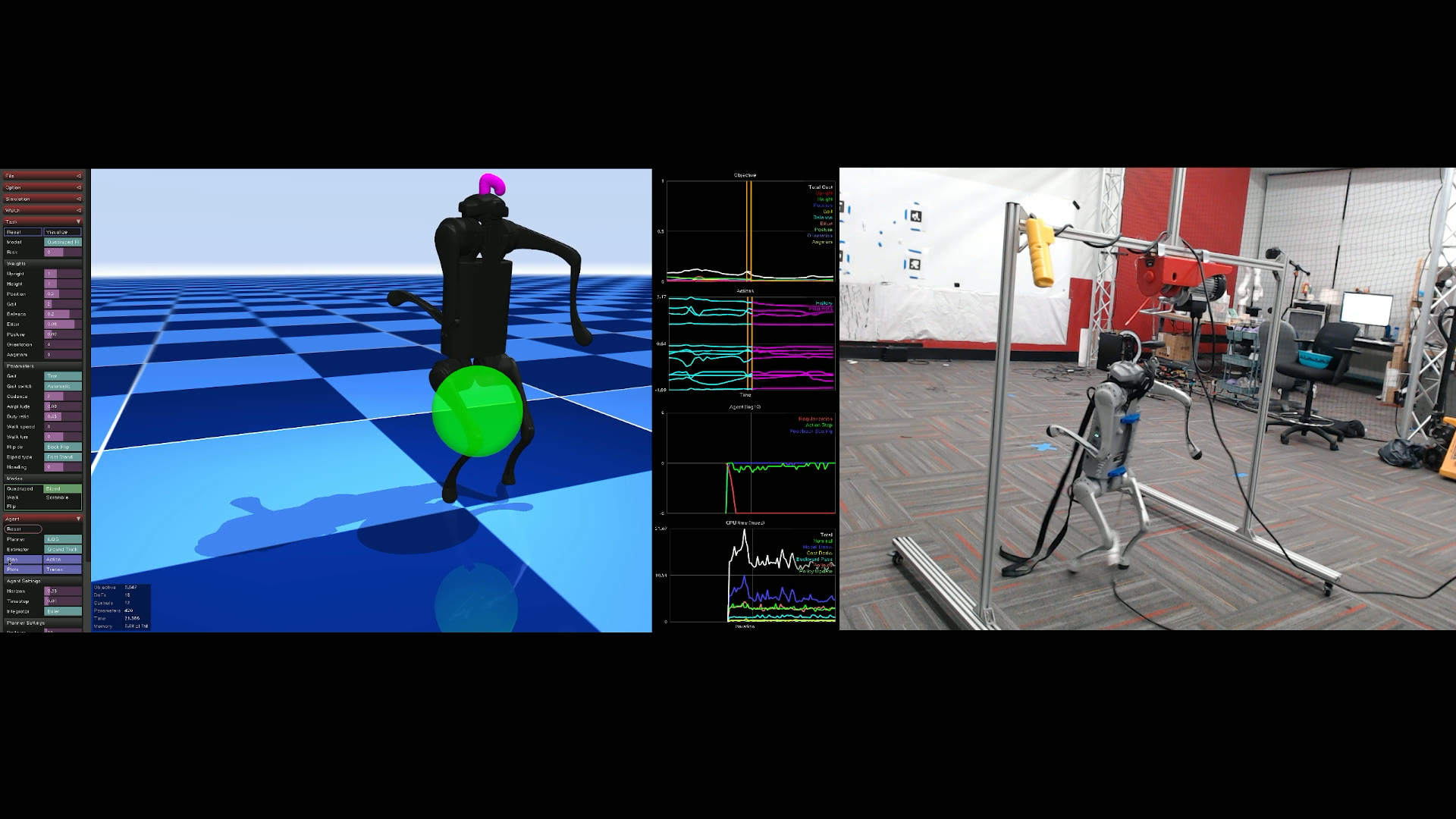}
    \hfill
    \includegraphics[trim={1350 300pt 300 400pt}, clip, width=0.24\linewidth]{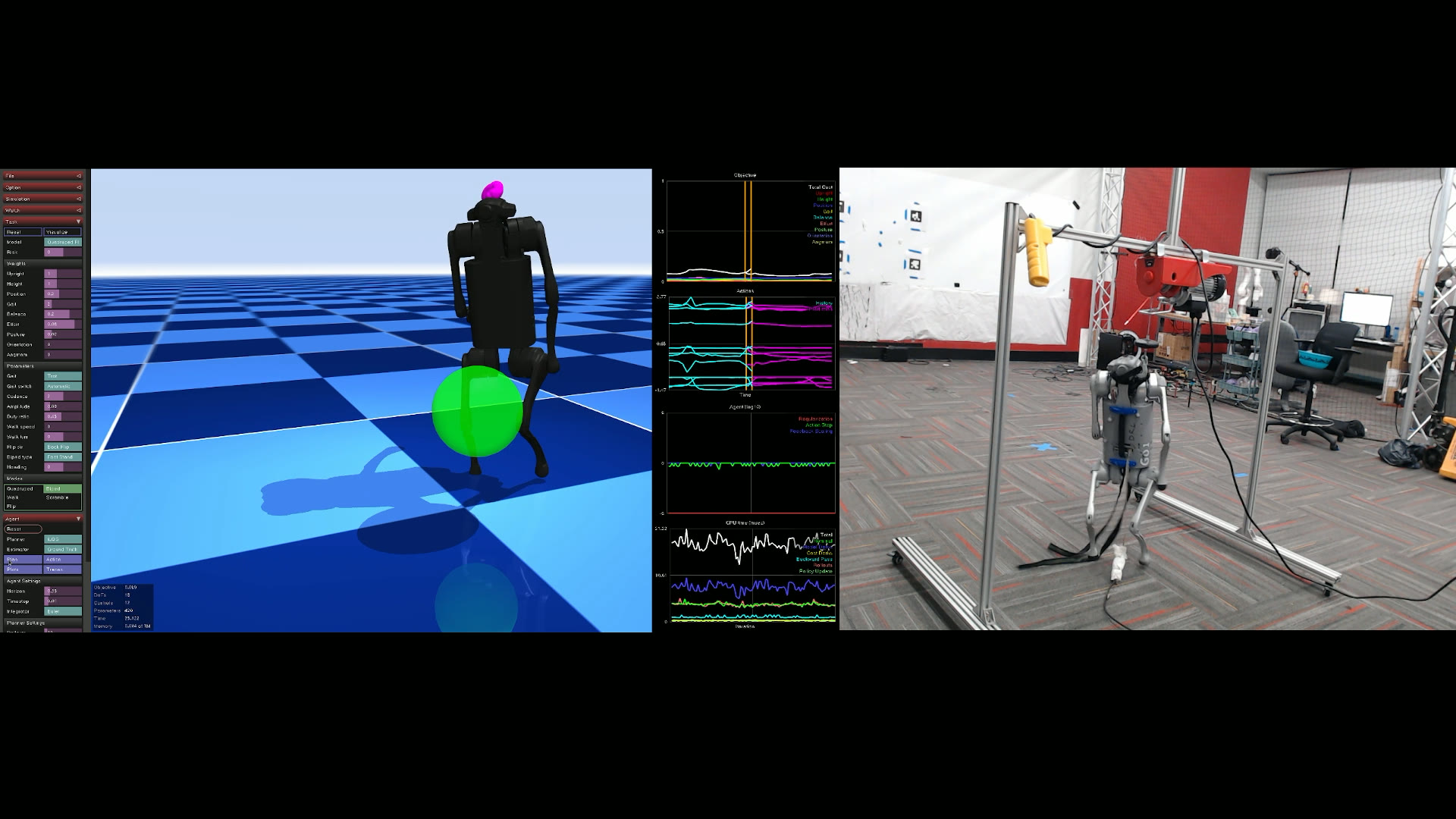}
    \hfill
    \includegraphics[trim={1350 300pt 300 400pt}, clip, width=0.24\linewidth]{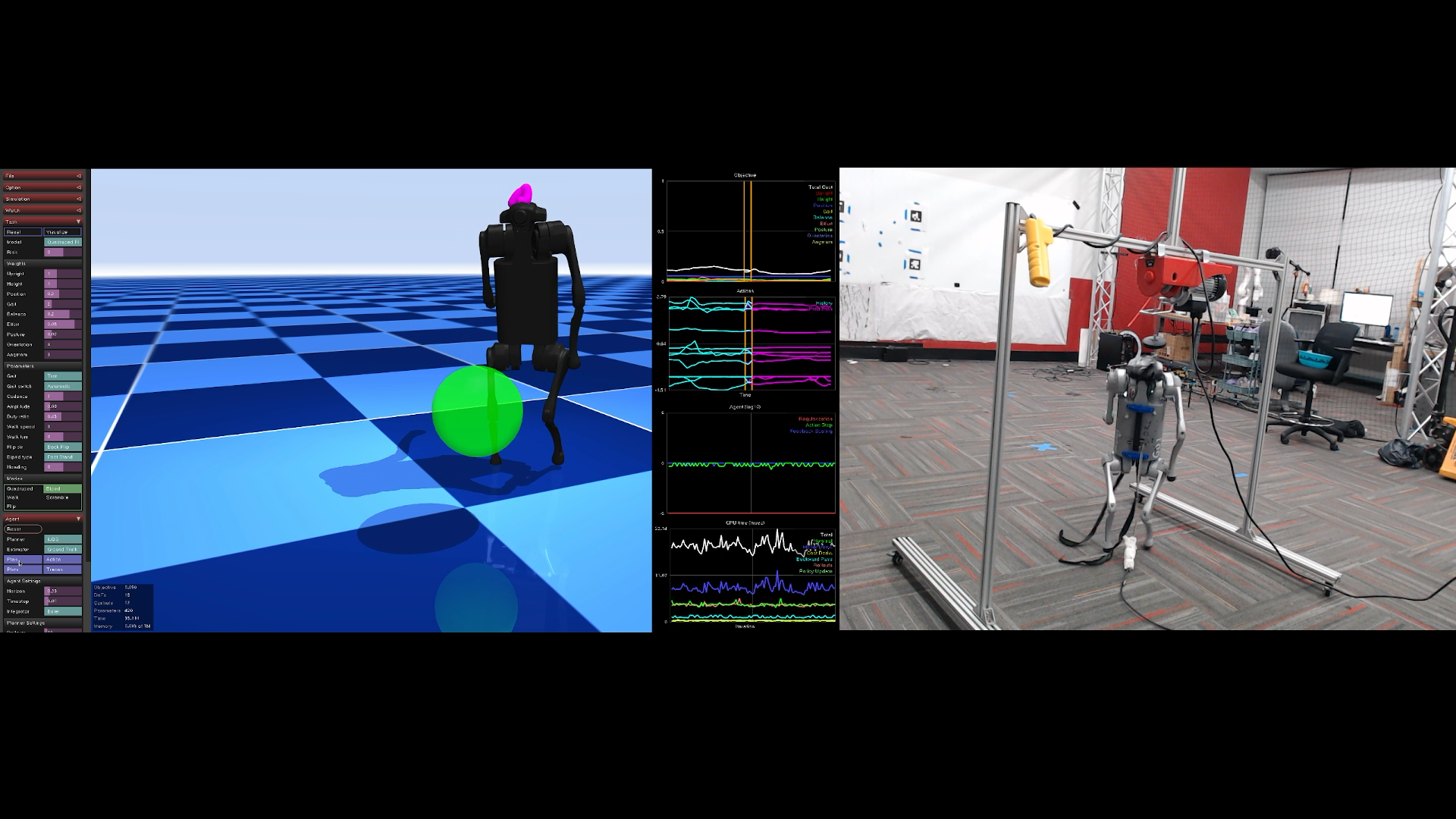}
    \caption{A Unitree Go1 quadruped robot transitions from quadruped to handstand mode (top row) and walking on its hind legs (bottom row) using the MuJoCo iLQR policy.}
    \vspace{-5mm}
    \label{fig:quadruped_biped}
\end{figure}

This paper aims to reduce this gap by providing an open-sourced baseline MPC algorithm and real-world legged robot implementation built on the MuJoCo physics engine~\cite{todorov2012mujoco}, a standard, easy-to-use open-source robotics simulator. We show that a standard gradient-based MPC algorithm, in particular the iterative LQR (iLQR) algorithm, based on MuJoCo is surprisingly capable of solving a variety of challenging \emph{real-world} tasks such as bipedal locomotion on a quadruped and full-sized humanoid robot in \emph{real time}. By leveraging the efficient C implementation of MuJoCo as the backend for the forward model and derivative computations, our real-time MPC approach reasons about both whole-body dynamics and collision detection in the model, something difficult to achieve in previous open-source whole-body MPC algorithms~\cite{mastalli20crocoddyl, alvarez2024realtime}. Additionally, we design an interactive GUI system that enables users to quickly modify key MPC parameters and observe \emph{real-world} robot behaviors alongside a simulated twin.
We hope that this effort lowers the barrier to entry for further model-based control research on legged robot hardware and eventually leads to accelerated research momentum. 
\begin{figure*}
    \centering
    \includegraphics[width=\linewidth]{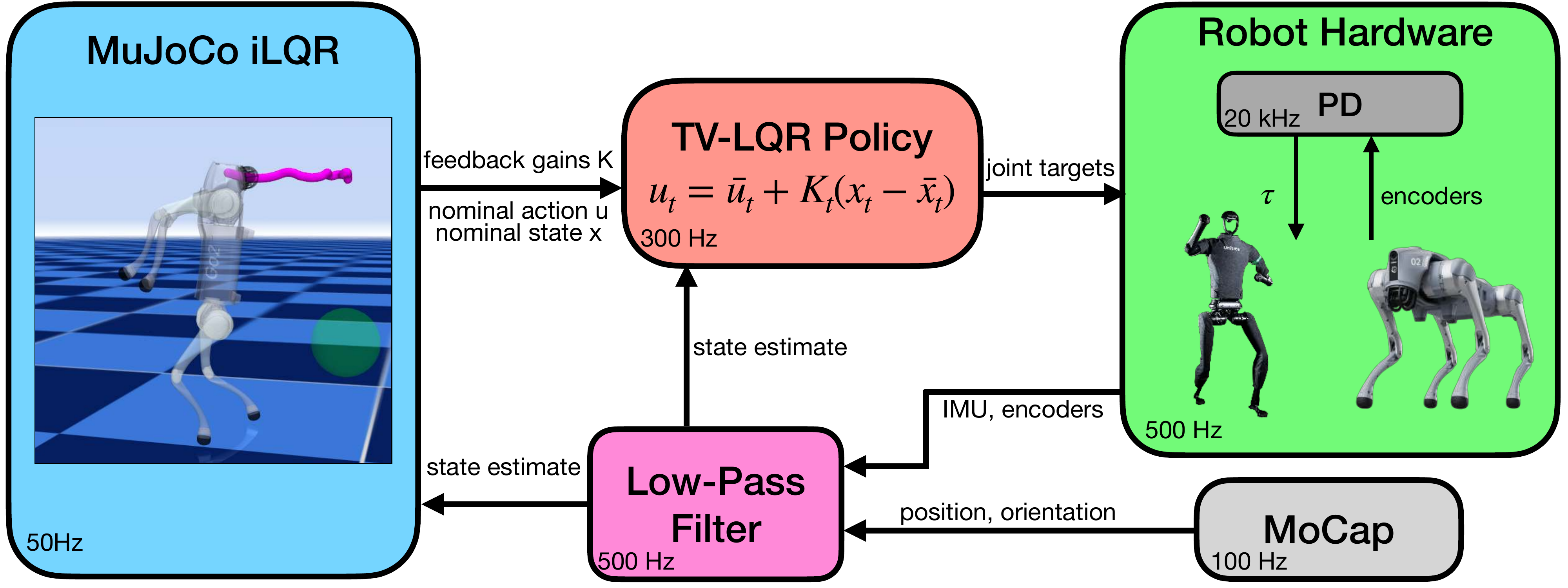}
    \caption{System diagram for deploying the MuJoCo iLQR policy to the Unitree Quadruped and Humanoid robots. The iLQR algorithm provides control, state, and time-varying LQR (TV-LQR) feedback gain trajectories at $50$ Hz. The TV-LQR feedback policy can then be updated at $300$ Hz and passed to a joint-level PD controller. The robot's state is estimated by fusing onboard joint encoders and motion capture data. Live state estimates are updated in the planner at $300-500$ Hz and visualized in the MuJoCo MPC GUI. The cost categories are designed offline but the relative weights, goal locations, and iLQR hyperparameters can be adjusted by the user interactively in real time through the GUI.}
    \label{fig:system_overview}
\end{figure*}

Our specific contributions in this paper are:
\begin{itemize}
    \item A simple-yet-surprisingly-effective baseline whole-body predictive control algorithm for real-world legged robot locomotion.
    \item An open-source interactive GUI system for real-world predictive control of legged robots.
    \item A set of hardware experiments demonstrating the effectiveness of the baseline algorithm on both quadruped and humanoid robots across a variety of tasks.
\end{itemize}

The remainder of this paper is organized as follows: We begin by reviewing relevant literature 
on iLQR, whole-body MPC, and open-source efforts for model-based control 
in Sec. \ref{sec:related_works}. Next, we briefly introduce the MuJoCo contact model and iLQR in Sec. \ref{sec:mujoco_ilqr}, followed by key considerations and implementation details for transferring iLQR policies to hardware in \ref{sec:implementation_details}. Then, we cover the hardware setup and experimental results in Sec. \ref{sec:experiments}. Finally, we conclude in Sec. \ref{sec:conclusions} by discussing current limitations of our system and directions for future work.

\section{background and related works}\label{sec:related_works}
This section provides a brief review of the iLQR algorithm, a survey of relevant literature on MPC for legged robots, and other notable efforts for improving tooling for model-based control.

\subsection{Iterative Linear-Quadratic Regulator}\label{sec:related_works_ilqr}
Differential Dynamic Programming (DDP), originally introduced in~\cite{JacobsonMayne1970DDP}, solves the following nonlinear trajectory optimization problem:
\begin{equation}\label{eq:ilqr}
\begin{aligned}
    \min_{u_{0:T-1}} & \quad \sum_{t=0}^{T-1} l(x_t, u_t) + l_f(x_T) \\
    \text{subject to} & \quad x_{t+1} = f(x_t, u_t) ,\\
    & \quad x_0 = x(0),
\end{aligned}
\end{equation}
by iteratively solving the an locally approximated problem with Dynamic Programming.
In addition to the nominal control sequence $u_{0:T-1}$, DDP also produces a time-varying linear feedback policy:
\begin{align}\label{eq: feedback}
    u_t = \Bar{u}_t + K_t(x_t - \Bar{x}_t) + \alpha k_t
\end{align}
where $\Bar{u}$ and $\Bar{x}$ are the current solution, $K_t$ is the feedback gain matrix at time index $t$, $k$ is an improvement to the current nominal control, and $\alpha$ is the line search step size.
This single-shooting formulation only optimizes over the controls $u_{0:T-1}$ and recovers the corresponding states $x_{0:T}$ by rollingout the discrete-time dynamics $x_{t+1}=f(x_t, u_t)$. $l(x_t, u_t)$ and $l_f(x_T)$ are the running and terminal costs, respectively. In each iteration, derivatives of cost and dynamics w.r.t to the control sequence are computed around the current solution points in a process called linearization to form a subproblem with quadratic cost and linear constraints. 
The term iterative LQR~\cite{Tassa2012Synthesis, LiTodorov2004ILQR}, or iLQR, generally refers to the Gauss-Newton approximations of the original DDP algorithm that is often more computationally efficient. The iLQR algorithm can also be modified to handle control limits~\cite{Tassa2014ControlLimited}, state constraints~\cite{Howell2019ALTRO, jallet2023proxddp}, and contact~\cite{kong2023hybrid}.

As a single-shooting algorithm, iLQR maintains dynamically feasible state trajectories without convergence requirements, is amenable to warm starting, and naturally handles unstable systems, since the rollouts are performed with a feedback policy, all of these features make it appealing as an online controller. However, this method generally assumes the dynamics are smooth and differentiable. For robots with contact, the dynamics are non-smooth and the derivatives become nontrivial to compute~\cite{howelllecleach2022, kong2024saltation}. Our empirical results show that the combination of the MuJoCo soft contact model and its finite difference derivative approximation is sufficient for iLQR and, somewhat surprisingly, transfers well to robot hardware despite obvious model mismatch.






\begin{figure*}
    \centering
    \includegraphics[width=0.98\linewidth]{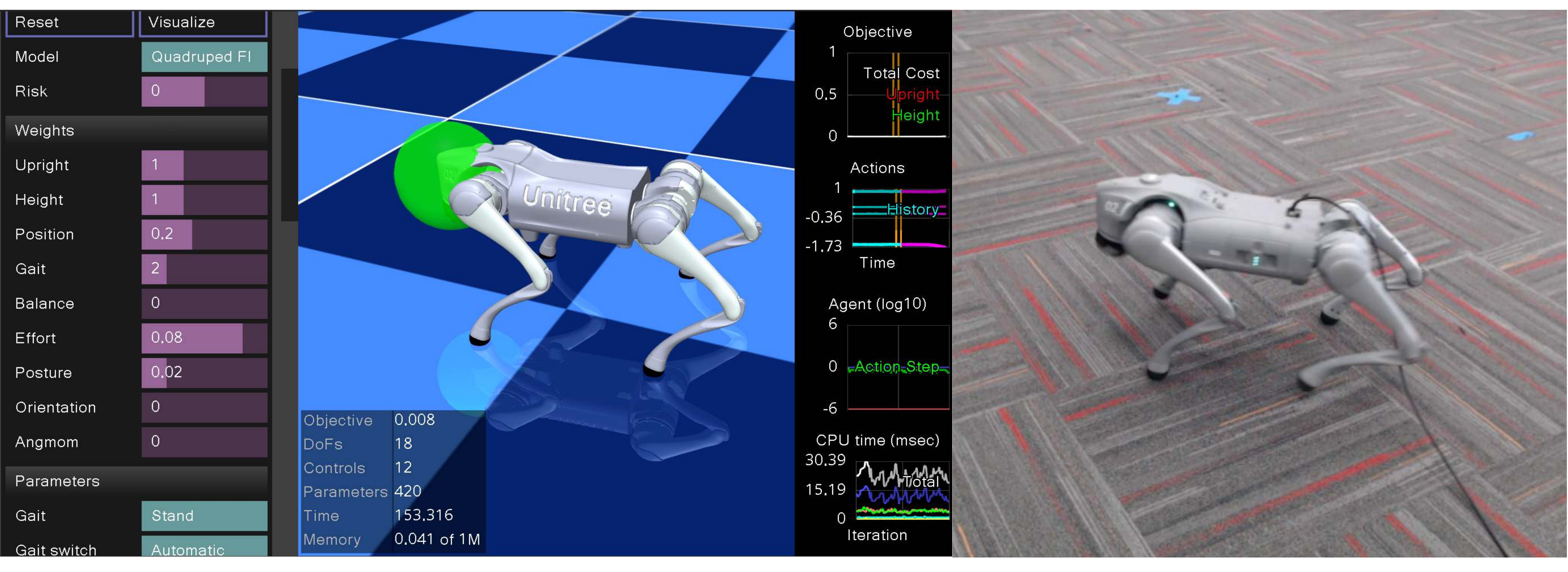}
    \caption{The MuJoCo MPC GUI for deploying legged robots on hardware. This GUI enables the user to interactively control the real-world robot by changing the target position defined as the green sphere. Additionally, the user can update the planner agent parameters, and observe the simulated states and the real-world robot behaviors in real time.}
    \label{fig:example_gui}
\end{figure*}


\subsection{Whole-body MPC for Legged Robots}
Whole-body nonlinear MPC presents challenges to legged robots primarily because of the real-time requirements of reasoning over possible contact modes and computation of the high-degree-of-freedom dynamics and its derivatives. Traditionally, real-time MPC has been achieved by simplifying models~\cite{cleach2024fast, bledt2018cheetah} and heuristically choosing the contact modes~\cite{bledt2018cheetah, raibert_dynamically_nodate}. Older works at whole-body MPC have been deployed on humanoid robots~\cite{koenemann2015whole}, they generally fall short of the real-time requirements as an online controller, limiting their real-world capabilities.
Thanks to advances in computer performance and increasingly mature implementations of dynamics libraries~\cite{carpentier2019pinocchio}, real-time whole-body MPC has become much more computationally trackable on quadruped~\cite{Kim_2024contact, lunardi2024reference} and humanoid robots~\cite{khazoom2024tailoring} in recent years. 
Another interesting thread of research explores enabling whole-body MPC through GPU parallelization~\cite{bishop2024reluqp}, but is so far limited to a single dynamics linearization of the model. 

Still, these methods generally require custom modeling of the robot dynamics, non-trivial analytical derivatives of the discontinuous contact dynamics, and custom optimization solvers to run in real time, making these prior works difficult to reproduce and iterate upon. This paper shows that a much simpler and more straightforward approach, modeling the robot using an off-the-shelf simulator and approximating the derivatives via finite differencing, can also be very effective for quadruped and humanoid locomotion \emph{without} specifying contact modes.



\subsection{Open-Source Tooling for Model-Based Control}
It is important to acknowledge previous open-source efforts to accelerate model-based control. In particular, MuJoCo MPC~\cite{howell2022predictive} implements a variety of derivative-based and derivative-free algorithms for predictive control and shows their effectiveness in simulation. \cite{alvarez2024realtime, xue2024fullordersamplingbasedmpctorquelevel} successfully demonstrates the sim-to-real transfer of sampling-based MPC using the MuJoCo dynamics on open-loop stable tasks like quadruped walking. Similar to~\cite{alvarez2024realtime}, this work builds on \cite{howell2022predictive} with a focus on sim-to-real of the derivative-based iterative LQR (iLQR) algorithm. Unlike sampling-based MPC, we show that iLQR can successfully tackle inherently open-loop unstable tasks such as bipedal walking.

Outside the MuJoCo ecosystem, Pinocchio~\cite{carpentier2019pinocchio} has become a popular toolbox in the community for efficiently computing rigid body dynamics and its derivatives. This dynamics toolbox has also enabled a variety of nonlinear trajectory optimization libraries designed for contact-rich robotics tasks like legged locomotion. For example, OCS2~\cite{OCS2} offers open-source software for quadruped locomotion with fixed contact modes, while Crocoddyle~\cite{mastalli20crocoddyl} and Aligator~\cite{aligatorweb} offer more versatile control for quadruped and humanoid robots. However, these works do not leverage off-the-shelf simulators widely used in the robotics community and have a steep learning curve, which limits their reach and impact. On the other hand, Drake~\cite{drake} provides advanced modeling and simulation features for accurate contact and friction dynamics for verification but is typically not fast enough for real-time control.

In comparison to prior projects (Table \ref{tab:comp}), our approach bridges the current gap in tooling for model-based robotic control by enabling general controllers for both quadruped and humanoid robots using a popular, off-the-shelf, and fast robotics simulator. An additional advantage for using a mature simulator is the readily available collision detection algorithms that we can leverage during contact-rich planning and control. Additionally, we provide an interactive GUI for real-time control that enables rapid developments of robot behaviors in the real world.
\begin{table}[h] \label{tab:comp}
    \centering
    \caption{A comparison model-based control tools through contact}

    \begin{tabular}{c c c c |c}
        \toprule
         &  \cite{drake} & \cite{mastalli20crocoddyl},\cite{aligatorweb} & \cite{OCS2}  & Ours\\
         \toprule 
         Off-the-shelf simulator & \cmark & \xmark & \xmark & \cmark\\
         Whole-body collision detection & \cmark & \xmark & \xmark & \cmark\\
         No fixed contact mode & N/A & \xmark & \xmark & \cmark \\
         Real-time control       & \xmark & \cmark & \cmark & \cmark\\
         Interactive GUI     & \xmark & \xmark & \xmark & \cmark\\
         \toprule
    \end{tabular}
    \label{tab:imitation_comp}
    \vspace{-5mm}
\end{table}

\section{Iterative LQR with MuJoCo}\label{sec:mujoco_ilqr}
This section provides a brief description of the MuJoCo soft contact model, the MuJoCo MPC toolbox, the details of the MuJoCo iLQR implementation, and sim-to-real considerations.

\subsection{MuJoCo Soft Contact Model and Derivatives}
The MuJoCo physics engine implements a soft contact model that is a convex approximation of the non-convex, discontinuous contact and friction models. While the interpenetration phenomena between objects (for example, the robot's foot and the floor) may be considered physically unrealistic, this convex formulation is fast, efficient, and provides a guaranteed solution, something difficult to do when solving non-convex problems. Additionally, in theory, the soft contact model offers smooth derivatives through contact. While the analytical derivatives are not yet provided, the finite different derivatives can be computed with little additional effort from the original time-stepping simulation problem.

To approximate the model derivatives, we use the forward difference method 
\begin{align}\label{eq:finite_diff}
    f'(x) \approx \frac{f(x + \epsilon) - f(x)}{\epsilon}
\end{align}
as it requires only one additional simulation evaluation per dimension compared to two in centered difference, where $f$ is an arbitrary function with input $x$ and $\epsilon$ is the finite different tolerance. 

\subsection{Derivative Computation with MuJoCo}

To solve a single iteration iLQR problem from Eq. \ref{eq:ilqr}, we take a second-order Taylor expansion and solve the resulting subproblem via Dynamic Programming~\cite{JacobsonMayne1970DDP, LiTodorov2004ILQR}. 

We define the cost function as the following:
\begin{align}\label{eq:cost}
    l(x, u) = \sum_i^M w_i \cdot \text{n}_i\Big(\text{r}_i(x, u)\Big),
\end{align}
where $r$ is a residual vector to be reduced when solving the problem, $n$ is the norm function that returns a non-negative scaler, and $w$ is non-negative scaler weight defining the importance of a residual term.

We compute the exact cost gradients:
\begin{align}
    \frac{\partial l}{\partial x} = \sum \limits_{i} w_i \cdot \frac{\partial \text{n}_i}{\partial \text{r}} \frac{\partial \text{r}_i}{\partial x} \\
    \frac{\partial l}{\partial u} = \sum \limits_{i} w_i \cdot \frac{\partial \text{n}_i}{\partial \text{r}} \frac{\partial \text{r}_i}{\partial u}
\end{align}
and approximate the cost hessians as:
\begin{align}
    \frac{\partial^2 l}{\partial x^2} \approx \sum \limits_{i} w_i \cdot \Big(\frac{\partial \text{r}_i}{\partial x}\Big)^T\frac{\partial^2 \text{n}_i}{\partial \text{r}^2} \frac{\partial \text{r}_i}{\partial x}\\
    \frac{\partial^2 l}{\partial u^2} \approx \sum \limits_{i} w_i \cdot \Big(\frac{\partial \text{r}_i}{\partial u}\Big)^T\frac{\partial^2 \text{n}_i}{\partial \text{r}^2} \frac{\partial \text{r}_i}{\partial u}\\
    \frac{\partial^2 l}{\partial x \, \partial u} \approx \sum \limits_{i} w_i \cdot \Big(\frac{\partial \text{r}_i}{\partial x}\Big)^T \frac{\partial^2 \text{n}_i}{\partial \text{r}^2} \frac{\partial \text{r}_i}{\partial u}
\end{align}
where the derivatives of the norm $\text{n}$ (i.e. $\frac{\partial \text{n}}{\partial \text{r}}$, $\frac{\partial^2 \text{n}}{\partial \text{r}^2}$) are computed analytically and the Jacobians of the residual $\text{r}$ (i.e. $ \frac{\partial \text{r}}{\partial x}$, $\frac{\partial \text{r}}{\partial u}$) are computed via finite difference, Eq. \ref{eq:finite_diff}.

We take the original nonlinear discrete-time dynamics:
\begin{align}
    x_{t+1} = f(x_t, u_t)
\end{align}
and linearize as:
\begin{align}
    \delta x_{k+1} = \frac{\partial f}{\partial x} \delta x_k + \frac{\partial f}{\partial u} \delta u_k
\end{align}
where dynamics Jacobians $\frac{\partial f}{\partial x}$ and $\frac{\partial f}{\partial u}$ are computed via finite difference, Eq. \ref{eq:finite_diff}.
Note the because residuals in Eq. \ref{eq:cost} are implemented as MuJoCo sensors, we can efficiently compute all the Jacobians $\frac{\partial f}{\partial x}$, $\frac{\partial f}{\partial u}$, $ \frac{\partial \text{r}}{\partial x}$, and $\frac{\partial \text{r}}{\partial u}$ via a \emph{single} call to the MuJoCo finite difference utilities function~\cite{todorov2012mujoco}.

Once all derivatives are computed, we solve the resulting problem using the Riccati-recursion that produces the updated nominal controls $u$ and a time-varying linear feedback policy $K$, Eq. \ref{eq: feedback}.
We perform this update once before returning the current-best nominal trajectories and feedback policy without convergence checks. We then use the previous solution to warm start a new iLQR iteration with the latest state estimation, Fig. \ref{fig:system_overview}. Additionally, our interactive GUI allows the user to update the residual terms such as target height, goal positions, etc and adjust the weights assigned to each residual term in real-time on the robot, Fig. \ref{fig:example_gui}.

\begin{figure}
   \centering
   \includegraphics[width=0.99\linewidth]{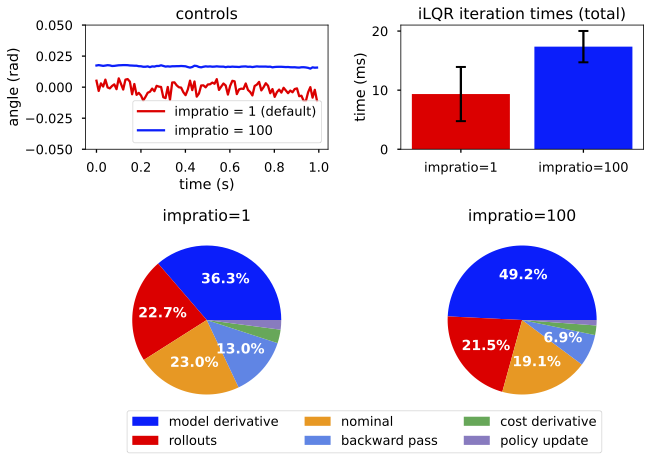}
   \caption{Top left: thigh joint control trajectories of two \texttt{impratio} contact settings on a quadruped robot standing in place. The default setting results in nonphysical foot slipping and jerky controls (red line) that are potentially dangerous on hardware. Top right: Increased \texttt{impratio} prevents this issue (blue line) but increases compute times (top right). $1$ standard deviation confidence interval for each bar is shown in black. Bottom: timing breakdowns of different iLQR components for each setting.}
   \label{fig:comp_impratio}
\end{figure}

\section{Implementation Details}\label{sec:implementation_details}
This section documents several implementation details for successfully deploying real-time MuJoCo iLQR on hardware. 
\subsection{Contact Modeling}
While the MuJoCo default contact parameters are fast and physically reasonable, they also often lead to contact slipping due to the solver's inability to enforce friction constraints~\cite{MuJoCoPreventingSlip}. This slipping dynamics from the planner model results in jerky control trajectories (Fig. \ref{fig:comp_impratio} top left) that are difficult to execute on the hardware.
We address this issue by increasing the \verb|impratio| value, which roughly corresponds to the solver's ability to trade off sliding versus penetration,  from the default $1$ to $100$. This change also has the added effect of increasing the solve times of the simulation problem and, as a result, the iLQR iteration times. For our quadruped system, the time for a single iLQR iteration increases from $\sim 10$ ms to $\sim 20$ ms on a $12$th-gen Intel i7 CPU. We do not find the added compute time to be an issue during real-world deployment. Interestingly, the moderate ground penetration from the softness of the contact model does not cause issues in the sim-to-real transfer in our experience.
\begin{figure}
   \centering
   \includegraphics[width=0.99\linewidth]{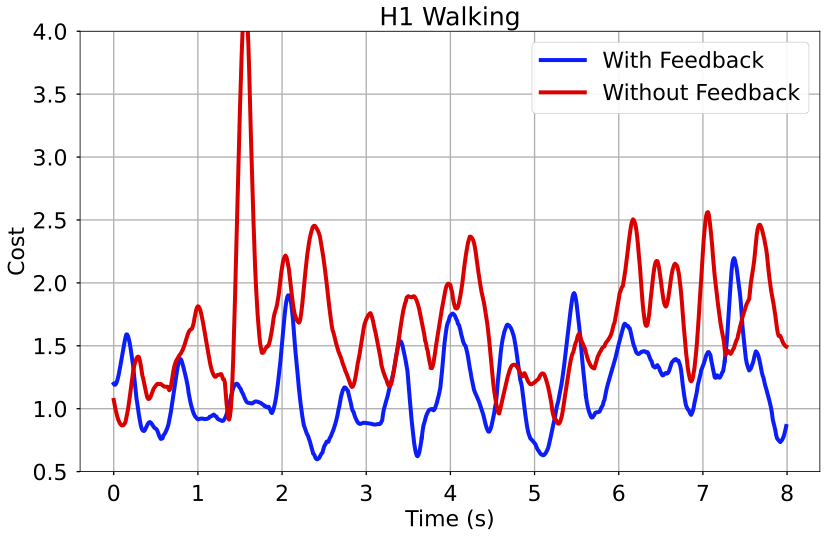}
   \caption{Cost comparison between the iLQR policy with (blue) and without (red) TV-LQR feedback gains applied to the nominal control sequence on an H1 humanoid robot trotting task on hardware. The TV-LQR policy improves task performance by $\sim 30\%$.}
   \label{fig:comp_feedback}
\end{figure}
\subsection{State and Action Representation}
We represent the state $x$ of the robot in the standard way, with a floating base position and attitude quaternion, followed by joint angles and corresponding linear, angular, and joint-angle velocities. We include a low-level joint-space PD controller in the dynamics model of the robot such that the inputs to the model that are optimized by iLQR are joint angle references. As a result, our approach achieves direct \emph{whole-body control} of the robot without the model hierarchies commonly seen in traditional model-based MPC algorithms~\cite{bledt2018cheetah}. By leveraging a fast off-the-shelf simulator, our method is simpler and more accessible than methods relying reduced-order models and hierarchical control approaches.

\subsection{Time-Varying Linear Feedback}
We compare the real-world performance of the iLQR policy with and without TV-LQR feedback on an H1 humanoid robot trotting-in-place task. The TV-LQR policy improves task performance but only \emph{marginally} compared to directly executing the nominal open-loop control sequence on the robot, resulting in an average of $30.1\%$ improved tracking performance over an $8$ second window, Fig. \ref{fig:comp_feedback}. Note that the cost spike around $1.5$s in the red line (without feedback) in Fig. \ref{fig:comp_feedback} indicates a temporary policy failure recovered with gantry support. The trial with feedback policy does not fail during the same window. Our empirical results are aligned with prior work~\cite{shirai2024linearfeedbacksmootheddynamics} which showed linear feedback on smoothed dynamics seems unsatisfactory for stabilizing contact-rich plans.
\subsection{Dynamics Derivative Approximation}
Computing the dynamics derivatives at every knot point is often the most computationally expensive step in iLQR, Fig. \ref{fig:comp_impratio}. We implement an optional heuristic by skipping dynamics derivative evaluations at some knot points along the planning horizon and instead interpolate from nearby evaluations. This heuristic is motivated by the fact that robot dynamics are approximately linear locally and, therefore, do not change much between nearby knot points~\cite{Han2016Data}. This is implemented as \verb|skip_deriv| in the GUI, where the integer value corresponds to the number of knot points to skip before computing the next derivative. In the tasks we consider in this paper, we find the iLQR update frequency to be sufficient when paired with TV-LQR policy. However, this option can still be beneficial for enabling iLQR in real time on more articulated systems or when compute is limited.

Note that while we choose to use forward difference approximation for computation reasons, we find that the centered difference method does not work on locomotion tasks, even in simulation without real-time requirements. We hypothesize that the derivative information into the contact (this can happen when perturbing the robot state in the negative height direction) is not informative during locomotion but leave further investigation for future work. 
\section{Experiments and Results}\label{sec:experiments}
This section presents the interactive GUI setup on robot hardware platforms and a variety of hardware experiments. We start by demonstrating our system on basic quadrupedal locomotion on Go1 and Go2 robots in Sec. \ref{sec:quadruped}. Next, we show that MuJoCo iLQR naturally extends to open-loop unstable tasks like quadruped walking on two legs in Sec. \ref{sec:quadruped-biped}. Finally, we deploy our system on a human-sized Unitree H1 humanoid robot in Sec. \ref{sec:humanoid}. Our open-source software and experiment videos are available at: \begin{center}
Ommitted for Anoynymous Review
\end{center}
\subsection{Interactive GUI for Real-World Legged Robots}
We interface with the MuJoCo MPC GUI and iLQR planner via Python for robot hardware deployment. The interface takes the latest state estimation and receives the current control. The state estimation is computed by fusing Optitrack MoCap~\cite{OptiTrack} position and attitude measurements at $100$Hz via ROS and the robot joint angle position and velocity measurements at $500$Hz. The floating base linear and angular velocities in the body frame are calculated by applying a low-pass filter to finite difference values of position and attitude measurements. The planner's actions are communicated to the robot via Unitree SDK for the Go1 robot and Unitree SDK 2 for the Go2 and H1 robots. All robot controls are represented as joint targets and tracked using Unitree internal low-level PD controllers. We compute the MPC policies on a desktop equipped with a $13$th generation Intel i$9$ CPU chip with $20$ cores, a different CPU from earlier in the paper.
We update the iLQR policy at $\sim50$Hz and use the TV-LQR policy to stabilize the robots between solves at $\sim300$ Hz. Unless otherwise noted, we use a default prediction horizon of $0.35$s and discretize the dynamics at $100$Hz in the iLQR planner.Note that, while we use a desktop-level CPU in our experiments, we see similar policy update frequencies from recent top-end laptop Arm-based CPUs such as the M-series chips from Apple, but leave deployment for future work.

Since the MPC planners are implemented in C++ and update asynchronously, using the Python interface does not affect the planning frequency. While we observe a $\sim 3$ms message passing overhead from the Python API, the planner is robust to this unmodeled delay for the tasks presented in this paper.

\subsection{Quadruped Locomotion}\label{sec:quadruped}
We first validate our system using a simple quadruped locomotion task. The task is designed for the robot to follow a reference gait and walk to the target locations. We successfully deploy the policy to the Unitree Go1 and Go2 robots with $12$ degrees of freedom (DoF) joint actuators. The GUI allows the user to move the target interactively and the real-world robot can follow the virtual target, as illustrated in Fig. \ref{fig:example_gui}. Note that while we keep the whole-body collision geometry in the planner model, generally only the $4$ spherical contact points are active during the locomotion tasks considered.

The residual terms for the locomotion task include tracking position and orientation while keeping balance and maintaining a desired torso height. Control efforts are penalized to minimize energy usage. We also include a nominal gait pattern in the residual, but note that this is not implemented as a constraint. As a result, the iLQR solver is free to discover new contact modes if they reduce the overall cost. More detailed descriptions of the residual terms can be found in Tab. \ref{tab: residual}. Bipedal locomotion and humanoid tasks follow a similar structure.

\subsection{Bipedal Locomotion for Quadruped Robots}\label{sec:quadruped-biped}
Next, we show that the iLQR policies naturally handles open-loop unstable tasks, such as a quadruped walking on two legs (Fig. \ref{fig:quadruped_biped}). In comparison, MPPI~\cite{alvarez2024realtime} can tackle locomotion tasks that are stable, but fails under unstable dynamics such as during bipedal walking. In Fig. \ref{fig:quadruped_biped}, we successfully enable a quadruped robot to walk gracefully on its back legs alone while using front legs to main balance and getting up to a hand stand pose from an initial quadruped configuration with the MuJoCo iLQR policy.

\begin{table}[h]\label{tab: residual}
\centering
\caption{Task Residual for Quadrupedal and Humanoid Locomotion}
\begin{tabular}{|l|p{6cm}|}
\hline
\textbf{Term} &  \textbf{Description} \\
\hline
Upright  & Keeps torso orientation upright (z-axis pointing up) \\
\hline
Height  & Controls torso height relative to average foot position \\
\hline
Position  & Tracks head position to target location (x, y, z) \\
\hline
Gait  & Controls foot lifting patterns during gait cycles (one per foot) \\
\hline
Balance & Keeps capture point within support polygon (x, y components) \\
\hline
Effort  & Penalizes actuator forces to minimize energy consumption \\
\hline
Posture  & Keeps joints near home configuration \\
\hline
Yaw  & Controls heading direction (x, y components of heading vector) \\
\hline
Angular & Controls rotational dynamics (x, y, z components) \\
\hline
\end{tabular}
\end{table}

\subsection{Humanoid Locomotion}\label{sec:humanoid}
\begin{figure}
   \centering
   \includegraphics[trim={200 170pt 950 250pt}, clip, width=0.24\linewidth]{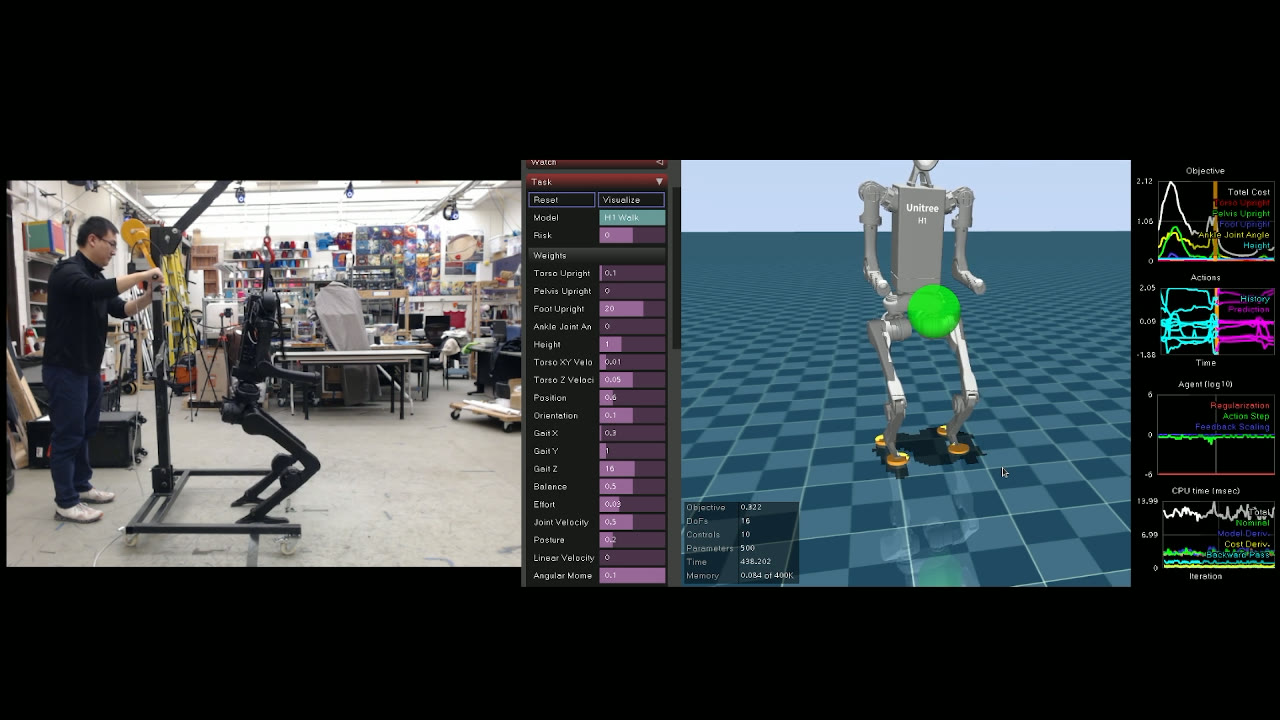}
    \hfill
    \includegraphics[trim={200 170pt 950 250pt}, clip, width=0.24\linewidth]{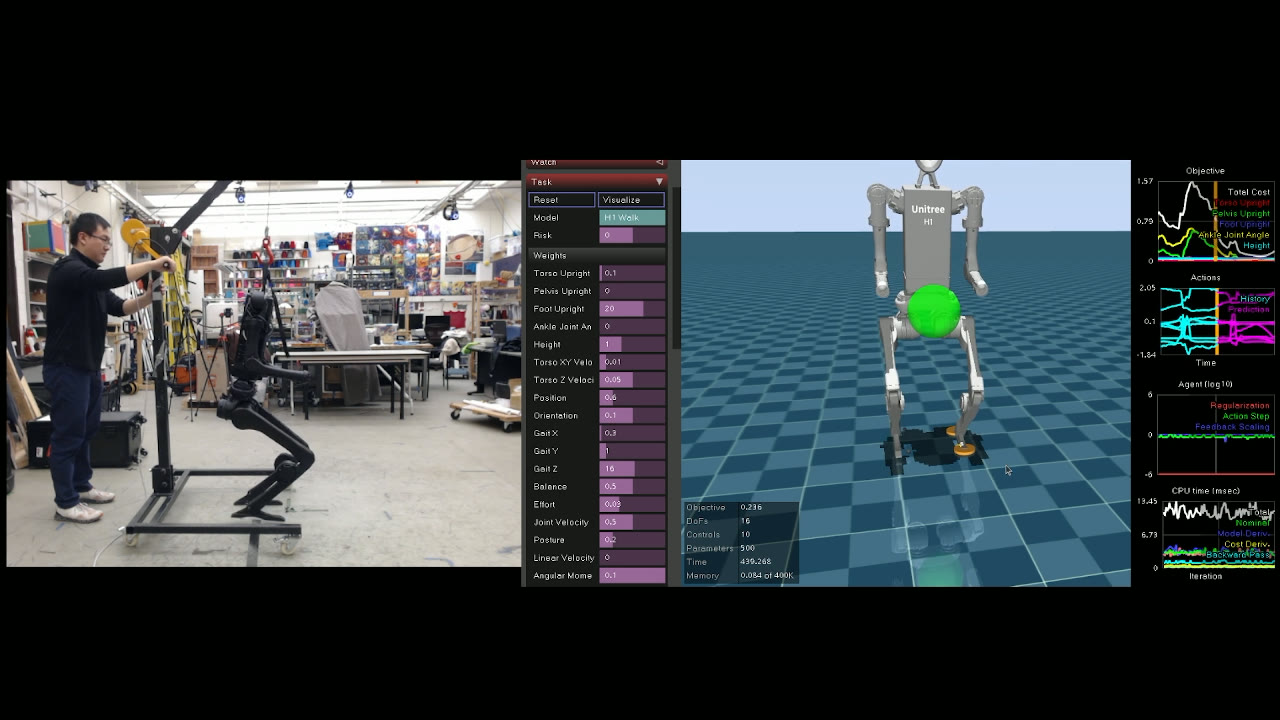}
    \hfill
    \includegraphics[trim={200 170pt 950 250pt}, clip, width=0.24\linewidth]{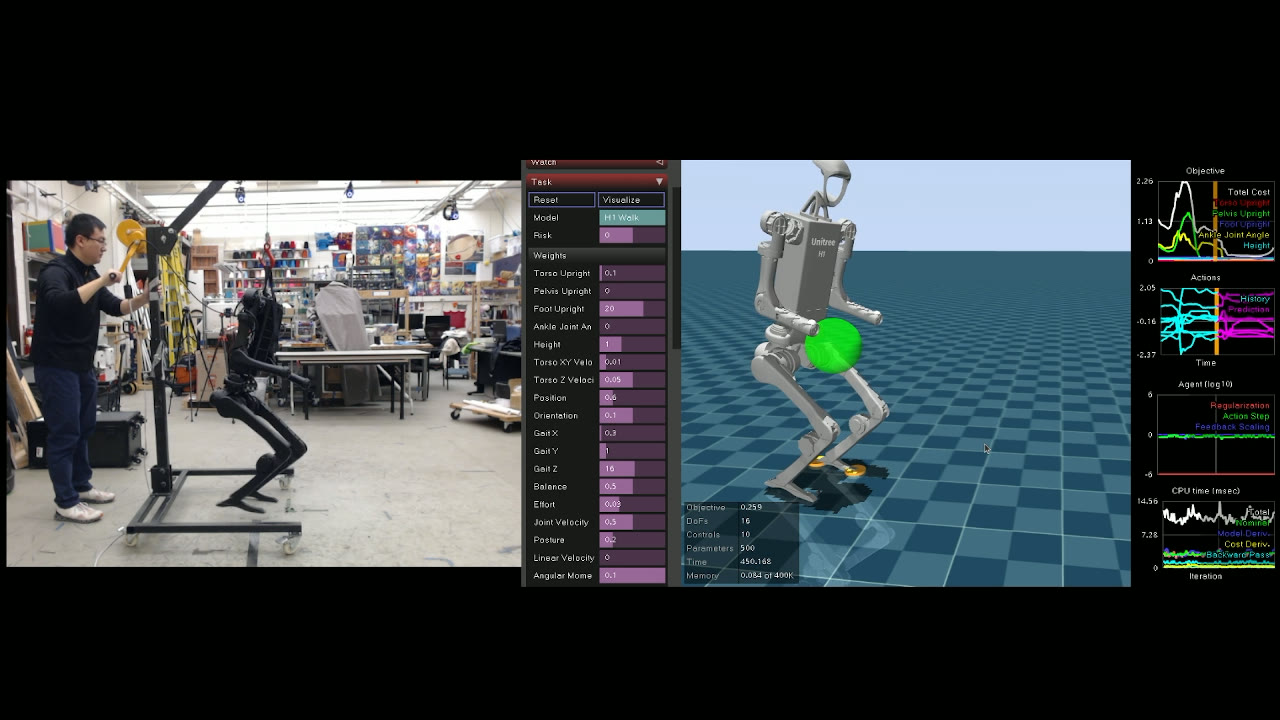}
    \hfill
    \includegraphics[trim={200 170pt 950 250pt}, clip, width=0.24\linewidth]{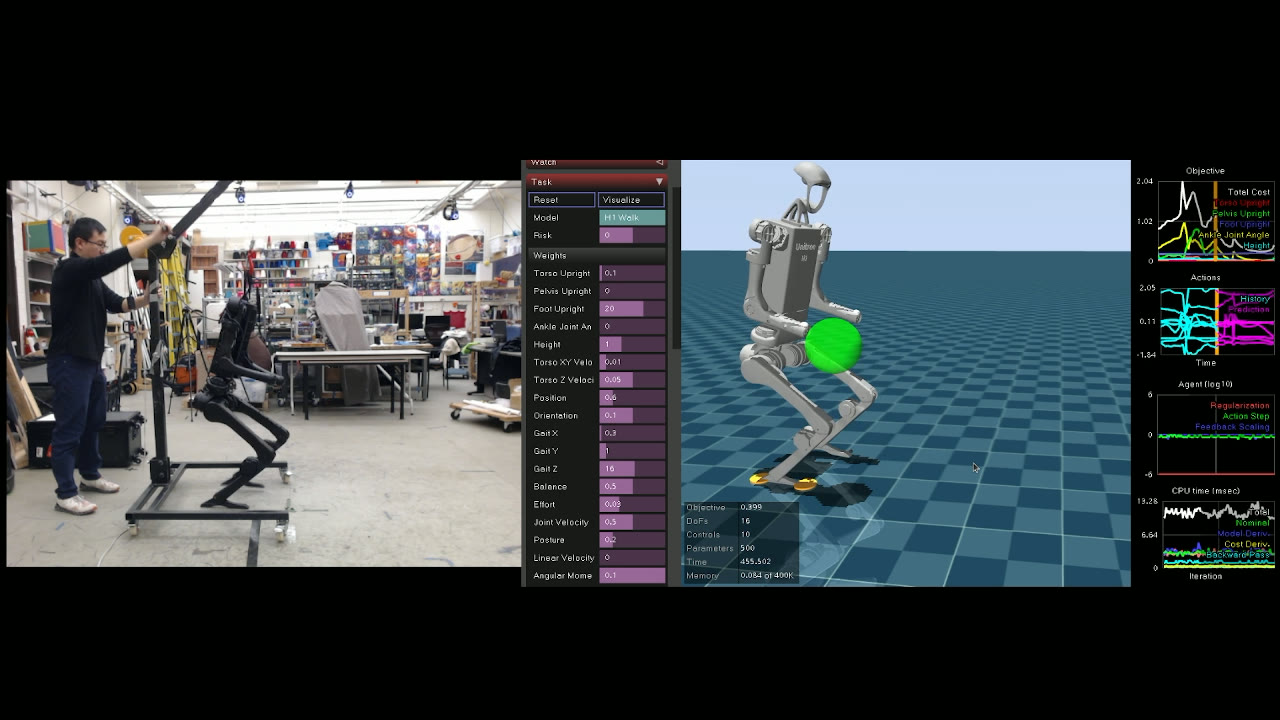}
   \caption{A Uniree H1 robot trotting in place with MuJoCo iLQR.}
   \label{fig:humanoid}
\end{figure}

Finally, we deploy the iLQR policy on an H1 full-sized humanoid robot to track a periodic trotting gait, as shown in Fig. \ref{fig:humanoid}. Similar to the quadruped locomotion task in Fig. \ref{fig:example_gui}, we ask the robot to walk to a position defined as the green sphere. To achieve real-time control on a more complex system, we disable collision checking in the robot body other than two spherical contact points on each foot. Additionally, we disable the DoFs in the robot upper body equipped with only relative encoders that provide unreliable joint angle estimates since they are not critical for locomotion.
 While this reduces the iLQR computation time, we do not believe it to be necessary for the success of the task. Overall, the system has $10$ DoF joint actuators and $4$ contact points. We use a prediction horizon of $0.5$s.

\section{Conclusions and Future Work}\label{sec:conclusions}
We present a very simple but surprisingly capable approach to whole-body MPC of quadrupeds and humanoid robots using MuJoCo. We hope that the successful hardware validation of this baseline method by leveraging a widely adopted physics engine can encourage researchers to leverage existing tooling for model-based control research. Despite the promising results, several key limitations remain for future work.

First, reliable state estimation remains a key challenge for model-based planning and control of legged robots. For this reason, many researchers prefer the RL paradigm which easily supports learning control policies directly from a history of sensor measurements. Our current system relies on marker-based motion capture to obtain good robot position measurements. Future work should develop easy-to-use tooling for full-state estimation from the robot's onboard sensors alone to enable our robots to walk outside controlled laboratory environments. Second, the community also needs additional tooling for rigorous system identification of the robot's joint actuation and contact dynamics to overcome the significant sim-to-real gap.

There are also several fundamental limitations of the iLQR algorithm. First, iLQR struggles with contact mode exploration. As a second-order derivative-based local planner, iLQR excels when the reference contact model schedule is provided in the task specification (cost function), which is typically the case for locomotion. However, for more contact-rich whole-body loco-manipulation tasks, derivative-free sampling-based methods have shown much more promise at discovering useful contact modes without prespecification~\cite {alvarez2024realtime}. Furthermore, iLQR rollouts and backward passes are both fundamentally serial operations. As computers become more parallelizable given the rise of multi-core CPUs and massively parallel GPUs, research on MPC algorithms that can leverage this computation paradigm becomes increasingly important.
Finally, iLQR suffers from other issues tied to its single-shooting nature, such as sensitivity to the initial guess, numerical instability, and poor convergence over longer horizons. Future work should extend the current MPC libraries to make multiple-shooting and collocation methods more accessible to the community.


\section*{ACKNOWLEDGMENT}
The authors thank Arun Bishop, Swaminathan Gurumurthy, and Shuo Yang for insightful discussions and feedback. This work is supported by a Google award.


\bibliographystyle{IEEEtran}
\bibliography{references}

\end{document}